\documentclass[runningheads]{llncs}
\usepackage[T1]{fontenc}
\usepackage{graphicx}
\usepackage{algorithm}
\usepackage{algpseudocode}

\begin{document}
\title{Self Similarity Matrix based CNN Filter Pruning}
%
%
\author{S Rakshith\orcidID{0000-0003-3658-5875} \and
	Jayesh Rajkumar Vachhani\orcidID{0000-0003-0267-4474} \and 
	Sourabh Vasant Gothe\orcidID{0000-0003-4737-2218} \and 
	Rishabh Khurana\orcidID{0000-0001-8225-4765}}
\authorrunning{S. Rakshith et al.}
%
\institute{Samsung R \& D Institute, Bangalore, India\\
\email{\{rakshith1.s, jay.vachhani,
	sourab.gothe, k.rishabh\}@samsung.com}}

\maketitle              
\begin{abstract}
Abstract. In recent years, most of the deep learning solutions are targeted to be deployed in mobile devices. This makes the need for development of lightweight models all the more imminent. Another solution is to optimize and prune regular deep learning models. In this paper, we tackle the problem of CNN model pruning with the help of Self-Similarity Matrix (SSM) computed from the 2D CNN filters. We propose two novel algorithms to rank and prune redundant filters which contribute similar activation maps to the output. One of the key features of our method is that there is no need of finetuning after training the model. Both the training and pruning process is completed simultaneously. We benchmark our method on two of the most popular CNN models - ResNet and VGG and record their performance on the CIFAR-10 dataset.

\end{abstract}
\section{Introduction}
State of the art neural networks that are intended to push the envelope in terms of accuracy and performance, have almost unlimited compute and space at their disposal. However, developers that are creating deep learning applications to be deployed on mobile devices do not have this luxury. With the rising use of AR, facial recognition and voice assistants, it has become more accessible to use ML solutions on mobile devices. There are multiple techniques which are employed to reduce the model footprint and accelerate its speed. These include matrix factorization \cite{dziugaite2015neural} \cite{ponti2020parameter}, knowledge distillation \cite{hinton2015distilling} \cite{cho2019efficacy} and model quantization \cite{hubara2017quantized} \cite{idelbayev2020flexible} \cite{jacob2018quantization} \cite{tailor2020degree}. There are two types of quantization techniques for reducing the model size - post training quantization and quantization-aware training. Quantization reduces the precision of the model parameters thereby reducing the model size and inference time.\\

One other way of tackling this issue is through network pruning. The process of pruning is accomplished by removing unimportant/redundant weights from a neural network thereby reducing its size. Pruning is a process which is beneficial for all deep learning models especially CNNs which have notoriously high computational constraints. We hypothesise that similar filters produce similar activations and contribute redundant information to the output layer, because of which such filters can be eliminated from the network. With the rise of mobile inference and machine learning capabilities, pruning becomes more relevant than ever before. Lightweight algorithms are the need of the hour, as more and more applications find use with neural networks.\\

There are numerous ways of selecting the weights to be pruned. In this paper, we aim to remove redundant weights from convolutional layers based on the similarity measure between filters of the same layer. In statistics and related fields, a similarity metric is a real-valued function that quantifies the similarity between two objects. By removing the similar filters, we can reduce the model size and computational overhead with negligible decrease in performance. In this paper, we devise new methods in computing and ranking redundancy among filters based on their mutual similarity values.

\begin{figure}
	\centering
	\includegraphics[width=0.7\linewidth]{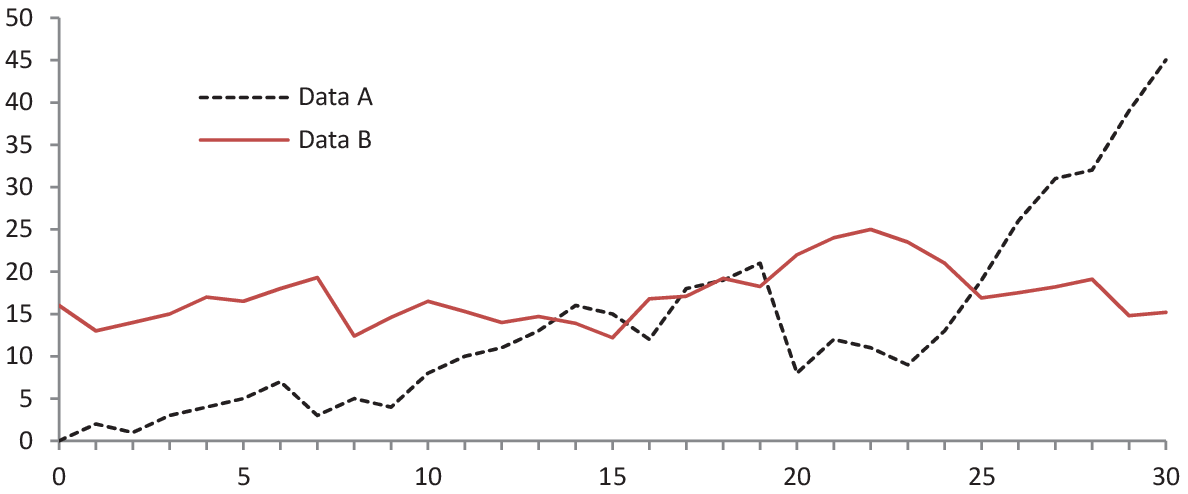}
	\caption{Pruning operation in 2D CNNs.}
	\label{fig:Pruning operation in 2D CNNs.}
\end{figure}

We choose to use the "Self Similarity Matrix" to capture the similarity measure. There are two steps in the process of shortlisting the most redundant filters for each layer. Firstly we compute the self similarity matrix (2D matrix) using an appropriate distance metric (L2, cosine, cityblock). Using the information captured in this matrix, we design two algorithms to identify pairs of similar filters and prune one of them. We wish to prove that with such a simple technique we are able to prune CNN models significantly, without much drop in accuracy.\\

Also our method is designed in such a way that we can prune the models while it is being trained without a need for finetuning. This saves valuable time and computational resources. We demonstrate our results on two state of the models ResNet and VGG on the CIFAR-10 dataset. By using the concept of SSM, we are treating the CNN filters as a sequence of multidimensional vectors. We propose two approaches - greedy approach and area based approach. Both have its own benefits and trade-offs.\\

Figure 1 highlights the process used for CNN filter pruning. Based on certain criteria according to the pruning algorithm, we identify rendundant filters and then prune them. After this another epoch of the training process is completed to update the weights of the remaining filters. After this the process is repeated iteratively until satisfactory performance is achieved. The details about each of the pruning algorithms are elaborated in the Proposed Method section.

\section{Related Works}
CNN pruning is one such field where there exists plenty of prior literature study. Many different techniques have been employed to rank and prune filters based on a criterion. In \cite{zhuo2018scsp}, the authors apply a novel spectral cluster method on filters and create efficient groups which are then used to choose redundant filters. The approach in \cite{lin2020hrank} is to prune filters based on the rank of their feature maps. Filters corresponding to lower rank feature maps are pruned as they are believed to contain less information. The authors in \cite{he2019filter} utilize the concept of geometric median to choose filters which are most similar to the remaining filters. For practical usecases, they use an approximate version of identifying the geometric median among the filters. In \cite{he2022filter}, in addition to the norm based criterion, the authors also use the geometric information among the filters.\\

\cite{singh2022passive} and \cite{li2020feature} are both pruning methods based on the similarity matrix. In \cite{singh2022passive}, the authors finetune the model post training, to obtain the final results. They show results for the accoustic scene classification on a very light weight network. In \cite{li2020feature} however, they employ two methods, "diversity-aware" and "similarityaware" to prune filters. There also many other novel approaches such as, \cite{yeom2021pruning} where authors mention a novel criterion for pruning inspired from neural network interpretability. In \cite{wang2021model}, they apply the pruning operation to filters of both convolutional as well as depthwise separable convolutional layers. In \cite{he2018amc}, The authors utilize a reinforcement learning algorithm to make the pruning process completely automated. In \cite{he2017channel}, pruning is based on a two-step process - Lasso regression and least square reconstruction. \cite{lin2019towards} utilizes a label-free generative adversarial learning to learn the pruned network with the sparse soft mask in an end to end manner and in \cite{ma2022pruning} the authors propose a variational Bayesian scheme for channelwise pruning of convolutional neural networks.

\section{Proposed Method}
In this section we elaborate on the entire procedure for pruning which is applied while the model is trained. One of the advantages of our approach is that we achieve good performance even without fine tuning of the model after pruning. This saves a lot of computational resources and time. We first begin by providing a brief introduction to the "Self-similarity matrix".

\subsection{Self-Similarity Matrix}
For two multidimensional vectors x and y, we can define a similarity function s : F×F → R to return a similarity score. The value s(x,y) is low if the vectors x and y are similar and large otherwise. For a given feature sequence X = ($x_1,x_2,...,x_N$), we can compute the N-square ’Self-Similarity Matrix’, $S \in R^{NXN}$ defined by,
\begin{equation}
	S(n,m)=s(x_{n},x_{m})
\end{equation}
Highly similar filters provide almost similar contribution to the output activation distribution. By removing one of the filters we can reduce model size with negligible decrease in performance. Here we devise new methods in computing and ranking filters based on their mutual similarity values. Self-similarity matrix is a graphical representation of similarity among a set of n-dimensional vectors. The basic idea can be seen in Figure 2, where we first identify sets of similar filters based on the values in the SSM. We then rank these filters and then prune them in the descending order of redundancy.

\begin{figure}
	\centering
	\includegraphics[width=0.7\linewidth]{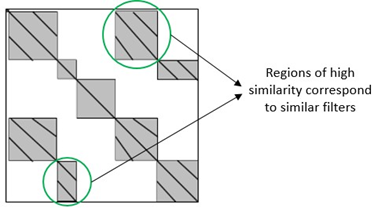}
	\caption{Self similarity matrix for CNN layer filters.}
	\label{fig:Self similarity matrix for CNN layer filters.}
\end{figure}
In Figure 3, we depict the overall pipeline of the pruning process. For each epoch of training, we iteratively prune the filters of each convolutional layer of the model. For a given convolutional layer, we flatten the corressponding set of 2D filters to create a list of multidimensional vectors. From these vectors we compute the SSM according to equation 1. Using this SSM as input, we can devise various methods for ranking and listing the redundant filters. We also input a constant "pruning ratio", which governs the number of filters to be pruned. After this we get a list of filters which need to be pruned from the convolutional layer.

\begin{figure}
	\centering
	\includegraphics[width=0.7\linewidth]{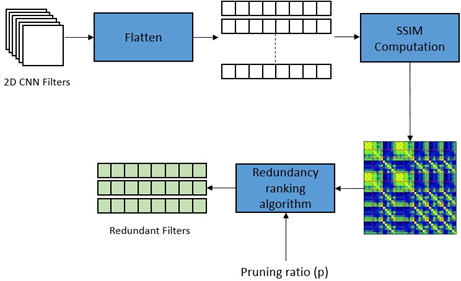}
	\caption{Highlevel description of the self similarity based pruning algorithm.}
	\label{fig:Highlevel description of the self similarity based pruning algorithm.}
\end{figure}
\subsection{Greedy Method}
In this method, we start by calculating the SSM for the filters of each of the convolutional layers. From the SSM, we select the filters to be pruned in a "greedy" approach. For the row of elements corresponding to each filter fi, we identify the filter fj corresponding to the smallest similarity value. We then add (fi,fj) along with the similarity value sij in a dictionary. This process is repeated for each row of the SSM. After this, we sort the elements based on the similarity values and start pruning one filter from the pair of filters. This can be visualized as a local method which considers the one-to-one similarity among only a pair of filters. The pseudo code is specified in Algorithm 1.

\begin{algorithm}
	\caption{Greedy Algorithm}\label{alg:cap}
	\begin{algorithmic}
		\State $S \gets Self Similarity Matrix (N X N)$
		\State $i\gets 0$
		\State $min\_arr \gets []$
		\While{$i \leq N$}
		\State $min\_arr[i]  \gets minimum(S[i,:])$
		\State $i \gets i+1 $
		\EndWhile
		\State $min\_indices \gets argsort(min\_arr)$
		\Return $min\_indices$
	\end{algorithmic}
\end{algorithm}
\begin{figure}
	\centering
	\includegraphics[width=0.7\linewidth]{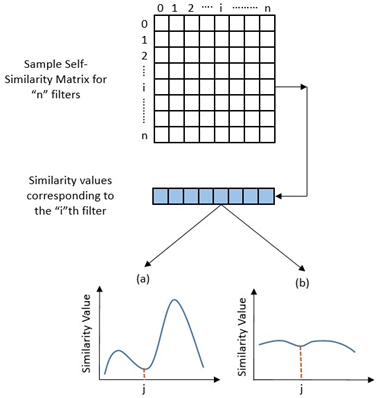}
	\caption{Comparison between area-based and greedy pruning algorithms.}
	\label{fig:Comparison between area-based and greedy pruning algorithms.}
\end{figure}
\subsection{Area based Method}
This is a more global method which considers the similarity of one filter with all the other filters and then ranks it based on that. Here we are treating the similarity values as a 1D curve and the area under this curve can be used as a global measure of similarity. The lower value of area corresponds to a filter which is more similar to the remaining filters. We choose to prune the most similar filters first. These are the filters which can be replaced without significant loss in performance.\\

In Figure 4, we have considered a sample SSM for a set of “n” filters. We also depict two possible plots corresponding to the “i” th filter. Even though the most similar filter corresponding to the “i” th filter is the “j” th filter in both the cases, we hypothesize that overall the plot in (b) is more similar to the rest of the filters. Based on the plot in (a) it is evident that even though “i” th and “j” th filters are very similar, the remaining filters are quite different from the “i” th filter, this would make pruning of the “i” th filter lose some accuracy of the model. To quantify this, we propose to calculate the area under the curve of the similarity values for each of the filter and then prune the filter whose plot corresponds to the lowest area. By doing this, we are capturing a more “global similarity” of each filter and making sure we are losing only the minimum amount of information by pruning the corresponding filter. The pseudocode is specified in Algorithm 2.

\begin{algorithm}
	\caption{Area Algorithm}\label{alg:cap}
	\begin{algorithmic}
		\State $S \gets Self Similarity Matrix (N X N)$
		\State $i\gets 0$
		\State $min\_arr \gets []$
		\While{$i \leq N$}
		\State $area[i]  \gets trapezoidal\_area(S[i,:])$
		\State $i \gets i+1 $
		\EndWhile
		\State $min\_indices \gets argsort(area)$
		\Return $min\_indices$
	\end{algorithmic}
\end{algorithm}

\section{Observation and Results}
In this section, we benchmark our pruning algorithms on two popular image classification networks namely, ResNet and VGG on the CIFAR-10 dataset. By benchmarking on such large CNN models, we can highlight the efficacy and usefulness of our method. We also compare the results with the method used in \cite{singh2022passive}. Currently we have pruned parameters from convolutional layers only, as they are the highest contributors for the FLOPs in the network. Therefore, all the results of pruned parameters are with respect to those corresponding to the convolutional layers. In Table 1 and Table 4, we list the original model performance details without pruning. For all the other results mentioned, we use a pruning ratio of 10\% each epoch for all the convolutional layers. We have used the ResNet-18 variant for the results with respect to the ResNet model.

\begin{figure}
	\centering
	\includegraphics[width=0.7\linewidth]{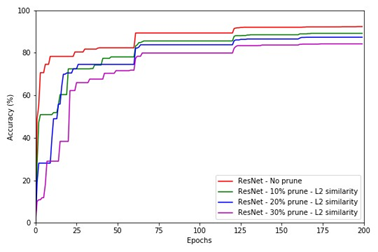}
	\caption{Accuracy variation for the ResNet model with different pruning ratios.}
	\label{fig:Accuracy variation for the ResNet model with different pruning ratios.}
\end{figure}

\subsection{ResNet Pruning}
\begin{table}
	\centering
	\caption{ResNet details}
	\begin{tabular}{|c|c|}
		\hline
		Total convolutional parameters & 2,30,832 \\
		\hline
		No prune accuracy (\%) & 92.78 \\
		\hline
	\end{tabular}
\end{table}

able 2 lists the accuracy drop of the ResNet model on the CIFAR-10 dataset after pruning for different methods. Table 3 shows the reduction in the convolutional layer parameters due to pruning. From Table 2, we see that even though there is a relative decrease (3\%) in accuracy for the area based method, there is close to 5x decrease in convolutional parameters compared to the CVSSP method. This shows the efficacy of considering the global similarity value of each convolutional filter. In case of the greedy method, both the drop in accuracy and convolutional parameters is high for 10\% pruning ratio. This can be adjusted accordingly to obtain requisite performance. In Figure 5, we have depicted the variation of accuracy as the ResNet model as it is trained for different pruning ratios. This figure clearly depicts the trade-off between the model size and accuracy.

\begin{table}
	\centering
	\caption{ResNet drop in \% accuracy after pruning on CIFAR-10 dataset (lower is
		better)}
		
	\begin{tabular}{|c|c|c|c|c|}
		\hline
		Pruning method & L2 distance & Cosine & Cityblock & KL Divergence \\
		\hline
		CVSSP \cite{singh2022passive} & 0.43 & 0.35 & 0.81 & 0.49 \\
		\hline
		Greedy Method & 3.81 & 14.37 & 4.02 & 0.14 \\
		\hline
		Area method & 0.51 & 3.61 & 0.32 & 3.85 \\
		\hline
	\end{tabular}
\end{table}

\begin{table}
	\centering
	\caption{ResNet reduction in \% convolution layer parameters after pruning on CIFAR-
		10 dataset (higher is better)
		.}
	\begin{tabular}{|c|c|c|c|c|}
		\hline
		Pruning method & L2 distance & Cosine & Cityblock & KL Divergence \\
		\hline
		CVSSP \cite{singh2022passive} & 11.02 & 7.17 & 11.02 & 2.26 \\
		\hline
		Greedy Method & 33.32 & 64.46 & 34.45 & 1.89 \\
		\hline
		Area method & 10.3 & 34.12 & 10.68 & 34.48 \\
		\hline
	\end{tabular}
\end{table}

adjusted accordingly to obtain requisite performance. In Figure 5, we have depicted the variation of accuracy as the ResNet model as it is trained for different pruning ratios. This figure clearly depicts the trade-off between the model size and accuracy.

\subsection{VGG Pruning}
Table 5 lists the performance of the VGG model on the CIFAR-10 dataset after pruning. Table 6 highlights the reduction in convolutional parameters in the VGG model due to pruning. From Table 5 and Table 6, it is evident that both the area based method and the CVSSP method provide similar results for VGG. This maybe due to the fact that the VGG model is almost 64x larger than the ResNet 18 model, and the advantage of the "global similarity" in the area method might be nullified for such a large scale increase in model size. We can also observe that all the distance metrics are not performing equally, especially KL-Divergence and cosine distance, exhibit poor performance.

\begin{table}
	\centering
	\caption{VGG details}
	\begin{tabular}{|c|c|}
		\hline
		Total convolutional parameters & 1,47,10,464 \\
		\hline
		No prune accuracy (\%) & 93.4 \\
		\hline
	\end{tabular}
\end{table}

\begin{table}
	\centering
	\caption{VGG drop in \% accuracy after pruning on CIFAR-10 dataset (lower is better)}
	\begin{tabular}{|c|c|c|c|c|}
		\hline
		Pruning method & L2 distance & Cosine & Cityblock & KL Divergence \\
		\hline
		CVSSP \cite{singh2022passive} & 0.2 & 9.2 & 0.4 & 10.2 \\
		\hline
		Greedy Method & 9 & 80.1 & 9.7 & -0.2 \\
		\hline
		Area method & 0.38 & 9.2 & 0.4 & 10.2 \\
		\hline
	\end{tabular}
\end{table}

\begin{table}[hbt!]
	\centering
	\caption{VGG reduction in \% convolution layer parameters after pruning on CIFAR-
		10 dataset (highrt is better)
	}
	\begin{tabular}{|c|c|c|c|c|}
		\hline
		Pruning method & L2 distance & Cosine & Cityblock & KL Divergence \\
		\hline
		CVSSP \cite{singh2022passive} & 19.91 & 48.85 & 19.19 & 48.88 \\
		\hline
		Greedy Method & 50.89 & 76.01 & 48.84 & 0.23 \\
		\hline
		Area method & 19.91 & 42.19 & 19.91 & 48.88 \\
		\hline
	\end{tabular}
\end{table}

\section{Conclusion}
In this paper, we have introduced two self similarity matrix based methods to identify and prune redundant filters from a CNN model. We highlight the tradeoffs for both the methods and also benchmark their performance on two state of the art CNN networks namely ResNet and VGG on the CIFAR-10 dataset. The greedy method for filter pruning depends on the local similarity among pairs of filters whereas the area based methods considers the global similarity among all the filters of a convolutional layer. The area based method offers better performance for relatively smaller models such as the ResNet 18, but performs similarly to that of the greedy method on larger models such as the VGG. As a future work, we wish to extend this approach beyond convolutional layers and make it applicable for all layers. We want to work towards making the pruning ratio as a hyperparameter which can be adjusted based on the model and dataset to provide a more customized pruning performance\cite{singh2022passive}.

{
	\bibliographystyle{splncs04}
	\bibliography{samplepaper}
}

\end{document}